\pdfoutput=1
\PassOptionsToPackage{prologue,dvipsnames}{xcolor}
\documentclass[11pt]{article}
\usepackage[dvipsnames]{xcolor}
\usepackage{ACL2023}

\usepackage{times}
\usepackage{latexsym}
\usepackage{framed}

\usepackage[T1]{fontenc}

\usepackage[utf8]{inputenc}

\usepackage{microtype}

\usepackage{inconsolata}
\usepackage{adjustbox}
\usepackage{graphicx}
\usepackage{subfig}
\usepackage{enumitem}
\usepackage{multirow}

\title{Summarization with Precise Length Control}

\author{Lesly Miculicich \quad Yujia Xie \quad Song Wang \quad Pengcheng He \\
  Microsoft \\
  \texttt{\{leslym,yujiaxie,song.wang,pengcheng.h\}@microsoft.com} \\}

\begin{document}
\maketitle
\begin{abstract}

Many applications of text generation such as summarization benefit from accurately controlling the text length. Existing approaches on length-controlled summarization either result in degraded performance or can only control the length approximately. In this work, we present a framework to generate summaries with precisely the specified number of tokens or sentences, while maintaining or even improving the text quality. In addition, we jointly train the models to predict the lengths, so our model can generate summaries with optimal length. We evaluate the proposed framework on the CNNDM dataset and show improved performance compared to existing methods.

\end{abstract}

\section{Introduction}
Controlling the length of the output is an important aspect of text summarization, as the desired length of the summary can vary depending on factors such as the size of the input document and the level of detail required in the summary. For example, a summary can range from a single sentence, providing a brief overview of the main topic or idea in the document, to several paragraphs providing a more detailed summary of the content. This can be particularly relevant in applications such as customizable summarization and constrained or fixed length summarization, where the text must fit specific device specifications such as screen width. 


The success of length-controlled summarization is measured by both the language quality and the length accuracy of the generated summary.
Initial approaches introduced the desired length as a parameter or vector embedding in the model \cite{kikuchi-etal-2016-controlling, liu-etal-2018-controlling} but these methods produced summaries with lower Rouge scores than their baselines. More successful approaches divide the training data into buckets or bins, each with specific length ranges (e.g. 0-30, 30-60) and use this information to build a summary prefix \cite{fan-etal-2018-controllable, he2020ctrlsum} or as a constrain \cite{takase-okazaki-2019-positional}. While the summary quality improves in comparison with previous approaches, this methods lose the ability for accurate length control with a specific number of tokens. Another way to control the summary length is by manipulating the probabilities of the end-of-sentence (EOS) token \cite{chan-etal-2021-controllable, liu-etal-2022-length},  but it may lead to fluency issues and lack of coverage when forcing an earlier sequence termination. 

In this work, we consider two practical cases of length-controlled summarization: token-level control and sentence-level control. We propose two easy-to-implement methods for the two cases, respectively.
The first, REverse Position Induced Length-cOntrolled Text generation \emph{REPILOT}, controls the precise number of tokens to produce. The second, explicit sentence enumeration \emph{SentEnum}, controls number of sentences. Both methods are highly accurate and show improvements on Rouges scores on two data-sets. Moreover, we jointly train the models to predict the optimal length of the summary given the input document. So our models can handle cases where the input length is not provided and it replaces the use length penalty during inference. Our evaluation results show that these methods produce summary quality that is comparable with state-of-the-art models and are significantly more accurate than previous approaches.

In summary, our contributions are: 1) A method for highly accurate length control of tokens and sentences with comparable or improved quality of the summary.
2) A baseline for sentence-based length control for summarization. 3) Length prediction in the summarization models to manage cases where the length is not given.

\section{Related Work and Baselines}
\label{sec:base}
Various methods for controlling the length of summaries have been proposed in the literature. \citet{kikuchi-etal-2016-controlling} proposed a method that uses a learnable length embedding input at the beginning of the decoding process, called \emph{LenInit}. They also experimented with inputting the remaining length at each time step of the decoding, called \emph{LenEmb}.
\citet{makino-etal-2019-global} improved upon this approach by optimizing the loss function with an overweight penalty, called \emph{GOLC}. \citet{liu-etal-2018-controlling} proposed a method that uses a length parameter as part of their CNN-based model, called \emph{LC}. 
\citet{liu-etal-2022-length} used two attention mechanisms, one for controlling the information selection and another for the end-of-sentence token. They first pre-trained a model with balanced length data \emph{LAAM}, and then fine-tuned it with original data \emph{PtLAAM}. \citet{saito2020length} proposed to control the summary length by inputting an extracted summary prototype \emph{LPAS}, and \citet{takase-okazaki-2019-positional} modified the sinusoidal positional embedding to allow length control during decoding.

Some additional studies have proposed generic approaches for controlled summarization with different features, including length. \citet{fan-etal-2018-controllable} proposed a method that divides the training data into buckets and prepends the corresponding bucket id to the summary. 
\citet{he2020ctrlsum} pre-trained a model by prepending extracted key-phrases from the summary. For length control they divided the training on predefined buckets, and used the mean key-phrases per bucket to control the summary length.
\citet{chan-etal-2021-controllable} proposed a method to control the decoding based on a Constrained Markov Decision Process. For length control, they use a cost function that computes the normalized distance between the bucket ids of the generated summary and the reference.

\section{Length Controlled Summarization}
Denote $x=\{x_1, \cdots, x_n\}$ as the input document, and $\ell$ as the desired length, i.e., number of tokes or sentences. Our goal is to train a probability model $p(y|x,l)$, where $y=\{y_1, \cdots, y_l\}$ is the summary.
If the length is not given, our models can also estimate the length of the summary as $p(l|x)$.


\begin{figure*}[h]
 \includegraphics[width=1\textwidth]{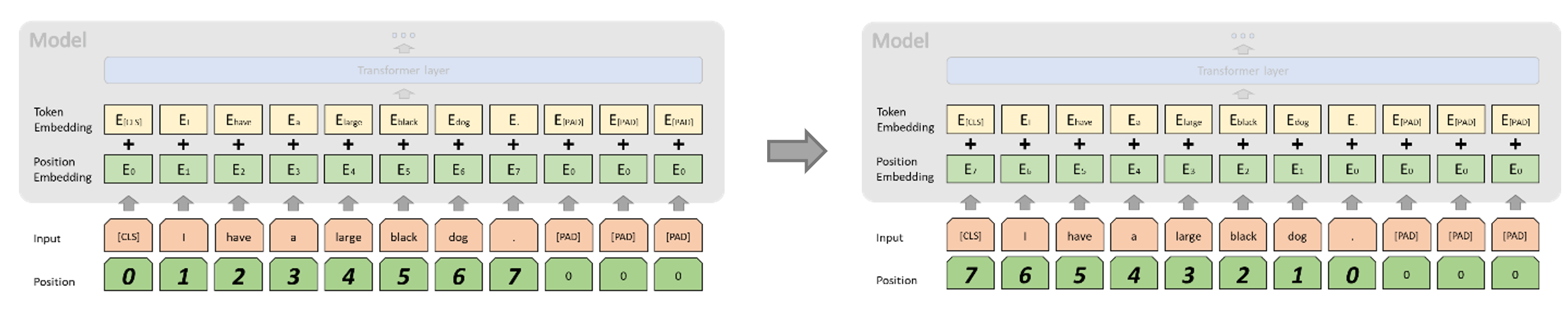}
\caption{Left: Regular model. Right: Reverse Position Induced Length-Controlled Text generation (REPILOT)}
\label{fig:repilot}
\end{figure*}

\subsection{REPILOT}
The REverse Position Induced Length-cOntrolled Text generation (REPILOT) method is a light-weighted solution for accurately controlling the generated text lengths. Specifically, we simply reverse the indices of position encoding as ``7, 6, …, 2, 1, 0'', as illustrated in the right of Figure~\ref{fig:repilot}. In this way, the model is aware of the information of how many more tokens should be decoded in each decoding step. By starting the position embedding with the target length, we can control the length.

In practice, we observe that training the model with the exact number of tokens may lead to abrupt ending of the generation, i.e., the generation will end once the position index reach 0, no matter whether the sentences have semantically ended. Therefore, we add a scalar noise to the position indices. Specifically, we sample a scalar $\delta \sim \mathcal{N}(0, 1)$, truncate it to integer, and add it to the sequence of position indices.

\subsection{SentEnum}
We propose a simple yet effective solution for generating a desired number of sentences in the output text. To the best of our knowledge, it is the first solution for controlling the number of sentences to generate; as previous  solutions control the number of tokens. Controlling sentences is a challenging problem because sentences boundaries like punctuation marks and spacing are often ambiguous. We guide the generation by using explicit enumeration of sentences. The required length is indicated as a number prefix to the summary. The inserted numbers are preceded by an special token \emph{[SN]} to differentiate them from the text\footnote{We preprocess the training data with the previously described annotation, using the sentence tokenizer from NLTK \url{https://www.nltk.org/api/nltk.tokenize.html}, and we post-processed the summary to remove the annotation at inference time.}, the following is an example:

\begin{framed}
\noindent \textcolor{RoyalBlue}{[SN]3} [SEP] \textcolor{WildStrawberry}{[SN]1} Nearly 40 endangered forest elephants were killed in 2 parks. \textcolor{WildStrawberry}{[SN]2} Sudanese poachers on horseback are believed to be responsible. \textcolor{WildStrawberry}{[SN]3} Forest and savanna elephant populations have declined drastically
\end{framed}


\subsection{Length Prediction}
The models are trained to predict both the length and the summary as multi-task learning. In the case of \emph{REPILOT}, we use a head classifier for predicting the number of tokens. 
The loss is computed with a weighted average from both length classifier and summarization, where we adopt the mean squared loss for the length prediction and the cross entropy loss for the text generation. In the case of \emph{SentEnum}, we simply train the model to predict the length prefix together with the summary as a text string. This method is easier to implement and preliminary experiments show similar performance as having a separate head to predict the number of sentences.

\section{Experimental Setup}

We initialize our models with Zcode++ \cite{arxiv.2208.09770}, a large pre-trained language model that reported strong results when fine-tuned in summarization. Unless specified we use the same hyper parameters and configuration. 

\noindent {\bf Data-sets:}
We perform experiments on CNNDM data \cite{nallapati-etal-2016-abstractive} and Arxiv \cite{cohan-etal-2018-discourse}.
Information and statistics are detailed in Appendix~\ref{sec:data}. 

\noindent {\bf Metrics:}
We use the standard Rouge score (R1, R2) for evaluating the summary outputs. To evaluate the success of the length control, we measure the accuracy (Acc.) and the mean absolute difference (Diff.). 


\section{Results and Analysis}
\subsection{Summary quality} 
Table~\ref{tab:cnndm} shows the Rouge scores on CNNDM. We compare with the reported results of previous works described in Section~\ref{sec:base}. All previous approaches use the length of the annotated reference summary to report rouge scores (marked as G in the Table~\ref{tab:cnndm}); and some of them use an external model to extract information from the document such as key-phases or -sentences (marked as E in the Table~\ref{tab:cnndm}) which are not directly comparable to ours. Our approach is the first to predict the expected length jointly with the summary. We show the results with our two models \emph{REPILOT} and \emph{SentEnum} with both golden and predicted lengths. In addition, we report the scores of Zcode++ finetuned on CNNDM. We use beam search of size 3 and no length penalty. The length penalty adjust the model to the typical length of the test set. We argue that the length prediction can replace the use of this hyper-parameter and preliminary experiments showed that our models archive better scores without it. We however report results of Zcode++ with and without length penalty. 

Both \emph{REPILOT} and \emph{SentEnum} using reference target lengths archive higher Rouge score than Zcode++. In addition, our models with predicted length performed better than Zcode++ without the length penalty adjustment, and on par with Zcode++ with length penalty. The results are also comparable with the reported results of similar approaches even though we do not use external model to extract information from the document to guide the summary generation.

\subsection{Length Control Accuracy on \emph{REPILOT}}
Table~\ref{tab:token_len_control} shows the performance of the length control measured in number of tokens for \emph{REPILOT}. We compare it with 3 baselines: a fine-tuned Zcode++ model and two models that group the summaries into different number of buckets and use the bucket ids as the length control \cite{fan-etal-2018-controllable, he2020ctrlsum}. The models are evaluated using golden lengths of reference summaries and using the same decoding method: greedy decoding without n-gram blocking. Our results show that \emph{REPILOT} achieved better ROUGE scores and lower mean absolute difference between requested and predicted lengths.

\subsection{Length Control Accuracy on \emph{SentEnum}}
Table~\ref{tab:sent_enum} shows the evaluation for \emph{SentEnum}. We compare it with three baselines: a fine-tuned Zcode++, a model that groups the summaries into \emph{Buckets} \cite{fan-etal-2018-controllable, he2020ctrlsum}, and a model that uses only the required number of sentences as prefix without sentence enumeration \emph{SentPrefix}. As the diversity of number of sentences in the summaries is small in CNNDM (Appendix~\ref{sec:data}), we include experiments in Arxiv. The results are obtained with greedy decoding without n-gram blocking which archived the best accuracy for all methods. We utilize the golden lengths as input, and evaluate the generated vs. the golden length. \emph{SentEnum}  is significantly more accurate to generate the required number of sentences and shows higher R2 scores. Appendix~\ref{sec:over_under} shows additional details about the percentage of over and under generation respect to the input length. \emph{SentEnum} shows the least percentage of errors. 

\begin{figure}
 \includegraphics[width=1\linewidth]{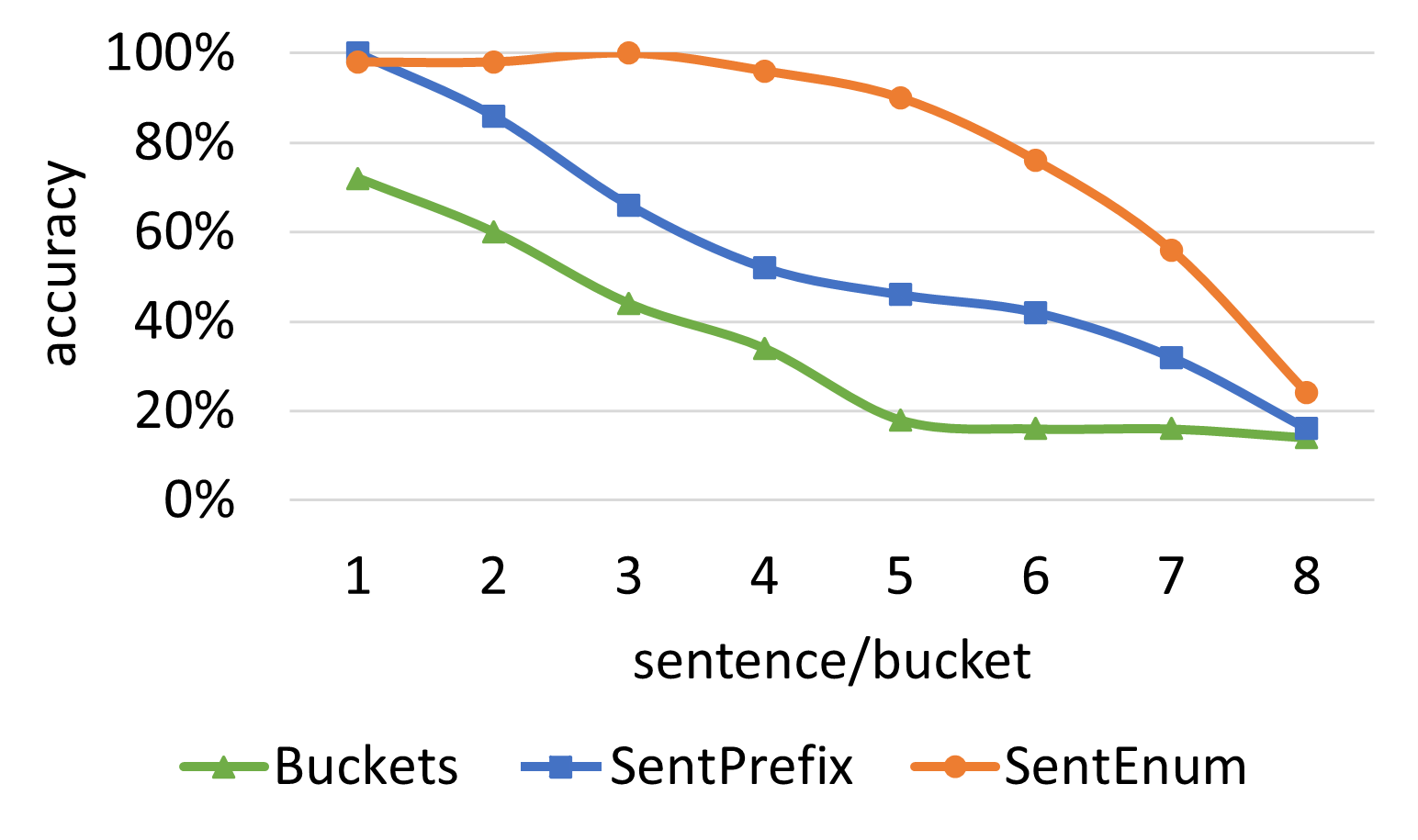}
\caption{Accuracy of the produced summary length respect to the input number of sentences. The evaluation is done on a out-of-distribution set.}
\label{fig:eval}
\end{figure}

Additionally, we evaluated the accuracy on an out-of-domain test set of 50 samples using length from 1 to 8 for all examples. Figure~\ref{fig:eval} shows the results comparing \emph{Buckets}, \emph{SentPrefix} and \emph{SentEnum}. The vertical axis show the accuracy of the generated length and the horizontal axis the input lengths. \emph{SentEnum} model shows higher accuracy for all input lengths, spatially in the middle range.  

\subsection{Length Prediction Accuracy}
Finally, we evaluate the length predictor. Table~\ref{tab:pred_len} shows the Diff. of the predicted  vs. the reference length. We compare with a \emph{Encoder-based classifier} trained using DeBerta. \cite{he2021deberta} with mean square loss. The jointly trained predictors are slightly more accurate than the individually trained ones showing that the multitask approach is effective.  

\section{Conclusions}
We study two simple methods for precisely controlling the length of tokens and sentences in text summarization. These techniques generated text with a specified length more accurately than previous methods. Additionally, we introduced a length predictor, making the models more versatile and easier to use without requiring an input length. These techniques can also be applied to other tasks such as text simplification and translation.

\begin{table}
    \centering
    \adjustbox{max width=\linewidth}{
    \begin{tabular}{l  c c c } 
    & & \textbf{R1}  & \textbf{R2} \\ \hline
    \emph{w/o pre-trained LM} \\ \hline
    LenInit \cite{kikuchi-etal-2016-controlling} & G &  25.87 & 8.27 \\
    LenEmb \cite{kikuchi-etal-2016-controlling} & G &  26.73 & 8.39 \\
    LC \cite{liu-etal-2018-controlling} & G & 35.45 & 14.50 \\
    GOLC \cite{makino-etal-2019-global} & G & 38.27 & 16.30 \\ 
    LenCtrl \cite{fan-etal-2018-controllable} & G & 39.16 & 15.54 \\
    LenAttn \cite{yu-etal-2021-lenatten} & G & 39.82 & 17.31 \\
    GPT2 CMDP \cite{chan-etal-2021-controllable}& G & 41.72 & 17.99 \\
    LPAS \cite{saito2020length} & GE & \emph{42.55} & \emph{20.09} \\ \hline
    \emph{w/ pre-trained LM} \\ \hline
    BART \cite{lewis-etal-2020-bart}& N & 44.16 & 21.28  \\
    BLPAS \cite{liu-etal-2022-length} & GE & \emph{42.95} & \emph{20.29} \\ 
    LAAM \cite{liu-etal-2022-length} & GE & \emph{43.55} & \emph{20.44} \\
    PtLAAM \cite{liu-etal-2022-length} & GE & \emph{44.17} & \emph{20.63} \\
    CtrlSum \cite{he2020ctrlsum}& E & \emph{45.65} & \emph{22.35} \\
    CtrlSum \cite{he2020ctrlsum}& GE & \emph{46.26} & \emph{22.60} \\ \hline
     Zcode++ \cite{arxiv.2208.09770} & N & 45.53 & 22.55 \\
     Zcode++ w/o length penalty & N & 45.19 & 22.41 \\
    + REPILOT & G  & \textbf{46.20} & 22.03 \\
    + REPILOT + length pred. & N & 45.61 & 22.13 \\
    + SentEnum.  & G  & 46.02 & \textbf{22.60} \\
    + SentEnum. + length pred. & N & 45.54 & 22.56 \\
   \end{tabular} }
	\caption{Evaluation Results on CNNDM data. G: Use length from the reference summary. E: Use extracted information from the document. N: None of the above.  }
	\label{tab:cnndm}

\end{table}

\begin{table}
    \centering
    \adjustbox{max width=0.75\linewidth}{
    \begin{tabular}{l c c c c c c}
     & \textbf{R1} $\uparrow$  & \textbf{R2} $\uparrow$  & \textbf{Diff.} $\downarrow$ \\ \hline
     Zcode++ & 44.76& 21.33 &		16.68  \\
     Buckets-10  & 45.82 &	21.76 &		5.84  \\
     Buckets-100  & 45.86 &	21.54 &		1.43  \\
     REPILOT & \textbf{46.36} & \textbf{22.08}  &\textbf{1.30}
    \end{tabular}}
     
    \caption{Results of the REPILOT modelon CNNDM using greedy decoding and without n-gram blocking.}
    \label{tab:token_len_control}

\end{table}

\begin{table}
    \centering
    \adjustbox{max width=.9\linewidth}{
    \begin{tabular}{l c c c c }
     & \textbf{R1} $\uparrow$  & \textbf{R2} $\uparrow$ & \textbf{Acc.} $\uparrow$ & \textbf{Diff.} $\downarrow$ \\ \hline
     \emph{CNNDM} & & & &  \\ \hline
     Zcode++ & 44.8& 21.3 &		60.1 &	0.5 \\
     Buckets & \textbf{45.8} &	21.8 & 87.1	& 0.1 \\
     SentPrefix & 45.7 & 21.7 &	94.0 &	0.1  \\
     SentEnum & 45.7 & \textbf{22.1} &	\textbf{98.6} &	\textbf{0.02}  \\ \hline
     \emph{Arxiv} & & & &  \\  \hline
     Zcode++ & 46.3	& 19.5 &	20.3 &	1.5  \\
     Buckets & \textbf{50.1} &	21.0 & 77.5 &	0.2  \\
     SentPrefix & 50.0 &	20.9 &  79.9 & 0.2 \\
     SentEnum & 49.8	& \textbf{21.3}	& \textbf{93.3}	& \textbf{0.1} \\
    \end{tabular}}
     
    \caption{Results of \emph{SenEnum} model using greedy decoding and without n-gram blocking.}
    \label{tab:sent_enum}

\end{table}

\begin{table}
    \centering
    \adjustbox{max width=\linewidth}{
    \begin{tabular}{l c c}
     & \textbf{Tokens} & \textbf{Sentences} \\ \hline
     Encoder-based classifier & 15.42  & 1.04   \\
     Jointly trained classifier  & 15.13 &   0.90  \\ 
    \end{tabular}}
     
    \caption{Diff. of the predicted length vs. gold length.}
    \label{tab:pred_len}
\end{table}

\section*{Limitations}
Our methods are not the perfect solution to control summarization granularity. We will keep exploring semantic aware length control to control the granularity of generated text in a more meaningful way. Additionally, the methods were not tested on unseen lengths in the training data. \emph{SentEnum} is less accurate when the input length is higher due to the fewer number of long examples in the training data. This suggest it doesn't generalize for all lengths.
Another limitation of \emph{SentEnum} is the quality of the sentence splitter. 
In our experiments, the errors introduced by the splitter did not have a significant impact on the results but it may not be the case for noisier data-sets or different languages.

\section*{Ethics Statement}
Our work is committed to comply with all applicable ACL Ethics Policy \footnote{\url{https://www.aclweb.org/portal/content/acl-code-ethics}}.
We presented methods which are simple to replicate and do not required high computation resources as they can use pre-trained language or summarization models. We acknowledge the risk of any text generation model to produce outputs that could lead to misinformation, bias or misuse. We however committed to use publicly available datasets whose content is relatively safe.

\bibliography{anthology,custom}
\bibliographystyle{acl_natbib}

\

\appendix

\section{Data Statistics}
\label{sec:data}

The following table shows the number of examples for training, development and testing for the reported data-sets:

\begin{table}[h]
    \centering
    \adjustbox{max width=\linewidth}{
    \begin{tabular}{l c c c}
    &  \textbf{Train} & \textbf{Dev.} & \textbf{Test}  \\ \hline
    CNNDM & 287,113 & 13,368 & 11,490  \\ 
    Arxiv & 202,914 & 6,436 & 6,440  \\

    \end{tabular}}
\end{table}

\noindent The following statistics are calculated from the summaries of the training sets.

\begin{table}[h]
    \centering
    \adjustbox{max width=\linewidth}{
    \begin{tabular}{l c|c|c|c|c|c|c}
     & \textbf{Max}& \textbf{Min} & \textbf{Mean} & \textbf{Med.} & \textbf{P75} & \textbf{P95}  & \textbf{STD}  \\ \hline
     \emph{CNNDM} \\ \hline
     Words & 1,246 & 4 & 49 & 46 & 57 & 85 & 20 \\ 
    Sent. & 36 & 1 & 3.8 & 4 & 4 & 6 & 1.3 \\ \hline
    \emph{Arxiv} \\ \hline
     Words & 26K & 2 & 278 & 164 & 237 & 482 & 587 \\
    Sent. & 20 & 1 & 6.1 & 6 & 8 & 10 & 2.2 \\ 
    \end{tabular}}

\end{table}

\section{Over and Under Length Generation}
\label{sec:over_under}

We count the percentage of examples with generated length shorter and longer than the gold length. We called \% Over) and \%Under generation.  This results were obtained with the \emph{SentEnum} model using with greedy decoding and without n-gram blocking in CNNDM and Arxiv data-sets. Given the data distribution, the number of shorter training examples is higher than the longer ones. Thus, all methods have tendency to under produce rather than over produce. However, \emph{SentEnum} show significantly better results.

\begin{table}[ht]
    \centering
    \adjustbox{max width=\linewidth}{
    \begin{tabular}{l  c c}
     &  \%\textbf{Over}$\downarrow$ & \%\textbf{Under} $\downarrow$\\ \hline
     \emph{CNNDM} & &  \\ \hline
     Zcode++ & 14.5 &	25.4 \\
     Buckets 	& 1.8	& 11.1 \\
     SentPrefix  &	1.1	& 4.9 \\
     SentEnum &	\textbf{0.8} &	\textbf{0.6} \\ \hline
     \emph{Arxiv} & &  \\  \hline
     Zcode++ &	19.9 &	59.8 \\
     Buckets &	6.8 &	16.6 \\
     SentPrefix & 5.9	& 14.7 \\
     SentEnum & \textbf{2.4} & \textbf{4.3} \\
    \end{tabular}}
\end{table}

\end{document}